\documentclass[conference]{IEEEtran}
\usepackage[utf8]{inputenc}
\usepackage[T1]{fontenc}
\IEEEoverridecommandlockouts
\usepackage[nocompress]{cite}
\usepackage{amsmath,amssymb,amsfonts}
\usepackage{algorithmic}
\usepackage{graphicx}
\usepackage{textcomp}
\usepackage{xcolor}
\usepackage[caption=false,font=footnotesize]{subfig}
\def\BibTeX{{\rm B\kern-.05em{\sc i\kern-.025em b}\kern-.08em
    T\kern-.1667em\lower.7ex\hbox{E}\kern-.125emX}}
\begin{document}

\title{A Lightweight Phenology-Aware YOLOv5 Framework for Tomato Growth Stage Detection in Resource-Constrained Bhutanese Greenhouse Environments\\ 
\thanks{This work was supported by the JSPS KAKENHI Grant-in-Aid for Scientific Research (B) (grant number JP25K03198 [S.N.]).\par
\vspace{3pt}This work has been submitted to the IEEE Xplore for possible publication. Copyright may be transferred without notice, after which this version may no longer be accessible.}}
\author{\IEEEauthorblockN{Sherab Gocha}
\IEEEauthorblockA{\textit{Graduate School of Information and Computer Science} \\
\textit{Chiba Institute of Technology}\\
Chiba, Japan \\
sgocha@plantech.gov.bt}
\and
\IEEEauthorblockN{Sou Nobukawa}
\IEEEauthorblockA{\textit{Graduate School of Information and Computer Science} \\
\textit{Chiba Institute Of Technology}\\
Chiba, Japan \\
nobukawa@it-chiba.jp}
}
\maketitle

\begin{abstract}
Accurate detection of tomato growth stages is essential for stage-specific greenhouse management and precision agriculture. In Bhutan, greenhouse cultivation is influenced by high altitude variability, pronounced diurnal temperature fluctuations, diffuse illumination, limited automation, and the scarcity of locally curated annotated datasets, which restricts the direct applicability of conventional deep learning–based detection models. To address these challenges, this work proposes Pheno-Lite + Efficient Channel Attention (ECA), a lightweight, phenology-aware object detection architecture derived from Ultralytics YOLOv5 and optimized for tomato growth stage recognition. A balanced dataset consisting of 2,464 annotated images was constructed using locally collected greenhouse images from Bhutan together with publicly available tomato plant images, which were further augmented to simulate Bhutan’s greenhouse environmental conditions. The dataset includes vegetative (820), flowering (824), fruiting (820), and background (26) samples to ensure robust class representation and improved background discrimination. Standardized preprocessing and controlled data augmentation strategies were applied to enhance generalization under realistic greenhouse conditions in Bhutan. The proposed architecture introduces two customized backbone modules: C3 PhenoLite, which improves fine-grained spatial and texture feature extraction through depthwise residual refinement, and C3 ECA, which incorporates efficient channel attention to strengthen inter channel feature interactions without dimensionality reduction. Experimental results demonstrate that the proposed model achieves 90.6\% precision, 88.8\% recall, and 92.6\% mAP@50, while maintaining a compact size of 4.0 million parameters and 10.9 GFLOPs at a 640 × 640 input resolution, highlighting its suitability for real-time, climate resilient greenhouse deployment in Bhutan.
\end{abstract}

\begin{IEEEkeywords}
Precision agriculture, tomato growth stage de- tection, lightweight object detection, mAP50, Precision, attention mechanisms, Preprocessing.
\end{IEEEkeywords}

\section{Introduction}
Smart agriculture represents a paradigm shift in modern food production, integrating artificial intelligence (AI), computer vision, the Internet of Things (IoT) systems, and automation to enable precision-based and data driven crop management \cite{Wolfert2017}. Within controlled environment agriculture (CEA), particularly greenhouse cultivation, intelligent systems facilitate adaptive regulation of irrigation, fertigation, ventilation, temperature, humidity, and light exposure \cite{Shamshiri2018}. Deep learning–based visual perception has become central to these systems, supporting automated plant phenotyping, yield estimation, disease detection, and growth stage recognition \cite{Kamilaris2018}. Among contemporary object detection frameworks, the Ultralytics-developed YOLO (You Only Look Once) family has gained widespread adoption due to its unified architecture, real-time inference capability, and favorable trade-off between detection accuracy and computational efficiency \cite{Jocher2023YOLOv8}. Such characteristics make YOLO-based models particularly suitable for embedded and resource-constrained agricultural deployments where continuous monitoring is required \cite{Badgujar2024}.
Recent advancements in agricultural computer vision have increasingly focused on multiclass phenological stage recognition rather than binary classification tasks. Large-scale phenology-oriented datasets for orchard crops have demonstrated the effectiveness of YOLOv8 and YOLOv10 based architectures in differentiating subtle developmental transitions under complex environmental conditions \cite{Wang2024YOLOv10}. In cereal crop monitoring, lightweight backbone integrations such as MobileNetV3 enhanced YOLO frameworks have achieved competitive performance in UAV-based rice growth stage recognition while maintaining reduced computational complexity \cite{Cai2024Rice}. Architectural refinements incorporating dynamic convolution and improved downsampling strategies have further enhanced discrimination between visually similar phenophases in wheat cultivation systems \cite{Chen2023Wheat}. In greenhouse contexts, improved YOLO-based detectors have been applied to multi-stage tomato detection under occlusion and non uniform illumination \cite{Liu2022Tomato}, while lightweight attention integrated frameworks such as GAE-YOLO have emphasized edge-oriented deployment for smart agriculture applications \cite{Liu2025GAEYOLO}.These studies collectively highlight three dominant trends, the development of phenology-specific annotated datasets, architectural refinement based on efficiency, and the integration of attention mechanisms to improve feature discrimination. However, most existing research has been conducted in technologically advanced agricultural ecosystems supported by large datasets and automated infrastructure, limiting their direct applicability to developing contexts characterized by environmental variability and resource constraints \cite{Liu2022Tomato}.
Bhutan presents a distinctive agro-ecological and infrastructural setting that underscores this research gap. Agriculture remains a critical socio-economic sector, yet greenhouse cultivation and automation are still at early stages of adoption \cite{Dorji2020}. The mountainous topography of the country results in significant fluctuations in diurnal temperature, seasonal humidity variability, and diffuse solar radiation during monsoon periods \cite{NCWC2019}. Greenhouse systems are primarily used for passive heat retention during winter rather than optimized year round precision cultivation \cite{MoAF2018}. Ventilation is largely natural, irrigation systems are often manual or based on simple drip mechanisms, and computational infrastructure in rural areas is limited \cite{Dorji2020}. Furthermore, there is no documented research addressing computer vision–based tomato growth stage detection in Bhutan, and locally curated phenological datasets are virtually nonexistent. These constraints necessitate a context-aware, lightweight detection framework capable of robust performance under variable environmental conditions and limited computational resources.
To address these challenges, this research develops a lightweight, phenology aware object detection framework tailored to Bhutanese greenhouse environments. A balanced dataset comprising 2,464 annotated images, 820 vegetative, 824 flowering, 820 fruiting, and 26 background samples was constructed to reflect realistic greenhouse variability. YOLOv5s was selected as the baseline architecture due to its compact parameterization, reduced computational complexity, and suitability for real-time deployment in resource constrained settings \cite{Jocher2022YOLOv5}. Building upon this baseline, the pro- posed Pheno-Lite + ECA architecture introduces a phenology-oriented lightweight refinement module (C3 PhenoLite) to enhance fine-grained spatial and texture representation, along with an Efficient Channel Attention (ECA) mechanism to adaptively recalibrate channel-wise feature importance without dimensionality reduction \cite{Wang2020ECANet}. Through systematic evaluation of detection accuracy, precision–recall characteristics, and computational efficiency, this study establishes a scalable and context-adaptive framework for stage-specific greenhouse monitoring, contributing toward improved productivity, resource optimization, and climate-resilient agricultural practices in Bhutan.

\section{MATERIALS AND METHODS}
\subsection{Dataset Acquisition and Development}
To address the absence of locally curated datasets for tomato growth stage detection in Bhutan, a Bhutan-specific image dataset was developed under real greenhouse operating conditions. Image acquisition was conducted using high-resolution smartphone cameras within greenhouse environments representative of typical cultivation practices in Bhutan. The images were captured from multiple view points, including frontal, lateral, and oblique perspectives, and at different times of the day to incorporate natural variations in illumination, shading, and plant orientation.\par
To enhance dataset diversity while maintaining contextual relevance, additional tomato images were selected from publicly available sources based on visual similarity in plant morphology, background complexity, and lighting characteristics. The final dataset comprised 2,464 annotated images, including 820 vegetative, 824 flowering, 820 fruiting, and 26 background-only images, ensuring balanced class representation and improved background discrimination.
\subsection{Dataset Organization and Annotation}
All images were manually annotated using bounding boxes in the YOLO format. Growth stage labels were assigned based on visible phenological characteristics, where the vegetative stage was defined by the presence of leaves only, the flowering stage by clearly visible open flowers, and the fruiting stage by the presence of discernible fruits. Background-only images were included to reduce false-positive detections in visually complex greenhouse scenes. The dataset was partitioned into training, validation, and testing subsets using an 82\%–12\%–6\% split, resulting in 2,030 training images, 290 validation images, and 144 testing images, after the data augmentation on training set with class balance preserved across all subsets.
\subsection{Data Preprocessing and Augmentation}
A standardized preprocessing and augmentation pipeline was applied to prepare the dataset for model training and evaluation. All images were resized to a fixed resolution of 640 × 640 pixels using a stretch-based resizing strategy compatible with the YOLO framework, ensuring consistent spatial dimensions and accurate alignment between images and corresponding bounding box annotations. This preprocessing step contributed to stable optimization during training and consistent evaluation across models.\par
To enhance model generalization under realistic greenhouse variability, data augmentation was applied exclusively to the training subset. Each training image generated up to four augmented variants using controlled geometric and photometric transformations. Geometric augmentations included horizontal flipping and random rotations within the range of $-10^{\circ}$ to $+10^{\circ}$. Photometric augmentations comprised grayscale conversion applied to 9\% of samples, hue variation within $\pm$5\%, saturation adjustment within $\pm$10\%, and brightness variation within $\pm$15\%. These augmentation strategies were designed to simulate diffuse illumination, partial shading, and viewpoint variations commonly observed in greenhouse environments while preserving phenological integrity.
\subsection{Baseline Model Selection and Training Setup}
YOLOv5s was selected as the baseline object detection architecture due to its lightweight design, favorable trade-off between detection accuracy and computational efficiency, and suitability for real-time deployment in resource-constrained agricultural environments. The model employs a convolutional backbone for feature extraction, a feature pyramid network–path aggregation network (FPN–PAN) neck for multi-scale feature fusion, and a multi-scale detection head, making it an effective and widely adopted baseline for agricultural computer vision applications.\par
Model training was conducted using the Ultralytics YOLOv5 framework. The network was initialized with pretrained weights to leverage transfer learning and accelerate convergence. All experiments employed identical dataset partitions, input resolutions, preprocessing pipelines, and augmentation strategies to ensure fair and reproducible evaluation. Default optimization parameters provided by the framework were used to maintain training stability and consistency across experiments.\par
All training and evaluation processes were performed on a high-performance computing workstation equipped with an NVIDIA A100-SXM4 GPU with 40 GB memory, providing sufficient computational capacity for efficient deep learning training. The system operated with NVIDIA driver version 580.82.07 and CUDA version 13.0, ensuring compatibility with the deep learning framework and stable execution of training workloads. Exclusive access to GPU resources was maintained throughout experimentation to ensure reproducibility of the reported results.The training configuration and computational environment used in this study are summarized in Table~\ref{tab:training_environment}.
\begin{table}[b]
\caption{Training Environment and Software Configuration}
\centering
\begin{tabular}{|l|l|}
\hline
\textbf{Component} & \textbf{Specification} \\
\hline
Operating System & Ubuntu 22.04 LTS (64-bit) \\
GPU & NVIDIA A100-SXM4 (40 GB) \\
NVIDIA Driver & 580.82.07 \\
CUDA Toolkit & CUDA 13.0 \\
Python Version & Python 3.12 \\
PyTorch & torch (CUDA-enabled) \\
TorchVision & torchvision \\
TorchAudio & torchaudio \\
NumPy & NumPy \\
OpenCV & opencv-python \\
Training Framework & Ultralytics YOLOv5 \\
\hline
\end{tabular}
\label{tab:training_environment}
\end{table}
\subsection{Evaluation Metrics and Experimental Protocol}
Model performance was evaluated using standard object detection metrics, including precision, recall, and mean Average Precision at an Intersection over Union threshold of 0.5 (mAP@50). Precision quantifies the proportion of correct positive detections, recall measures the model’s ability to identify all relevant objects, and mAP@50 reflects overall detection accuracy by aggregating precision–recall performance across all growth stage classes.\par
To ensure reproducibility and unbiased evaluation, all experiments were conducted using fixed dataset partitions, identical preprocessing and augmentation pipelines, and consistent input resolutions. Training, validation, and testing sets were strictly separated, and no test samples were used during model optimization or hyperparameter tuning. All models were evaluated under the same experimental conditions and metrics, enabling a fair and controlled comparison of detection performance.

\subsection{Proposed Pheno-Lite + ECA Architecture}
You Only Look Once (YOLO) is a one-stage object detection framework that performs object localization and classification in a single forward pass, enabling real-time detection. The architecture consists of a backbone for feature extraction, a neck for multi-scale feature aggregation, and a detection head for bounding box and class prediction. YOLOv5 employs a convolutional backbone with CSP-based modules such as the C3 block to enhance feature representation while maintaining computational efficiency. The C3 block is a cross-stage partial (CSP) bottleneck module composed of stacked convolutional layers with partial feature reuse, improving feature representation while reducing redundant computation. The neck fuses multi-scale features, and the detection head generates predictions at multiple scales, achieving a balance between detection accuracy and inference speed.\par
\begin{figure}[t]
\centering
\includegraphics[width=\columnwidth,height=0.35\textheight,keepaspectratio]{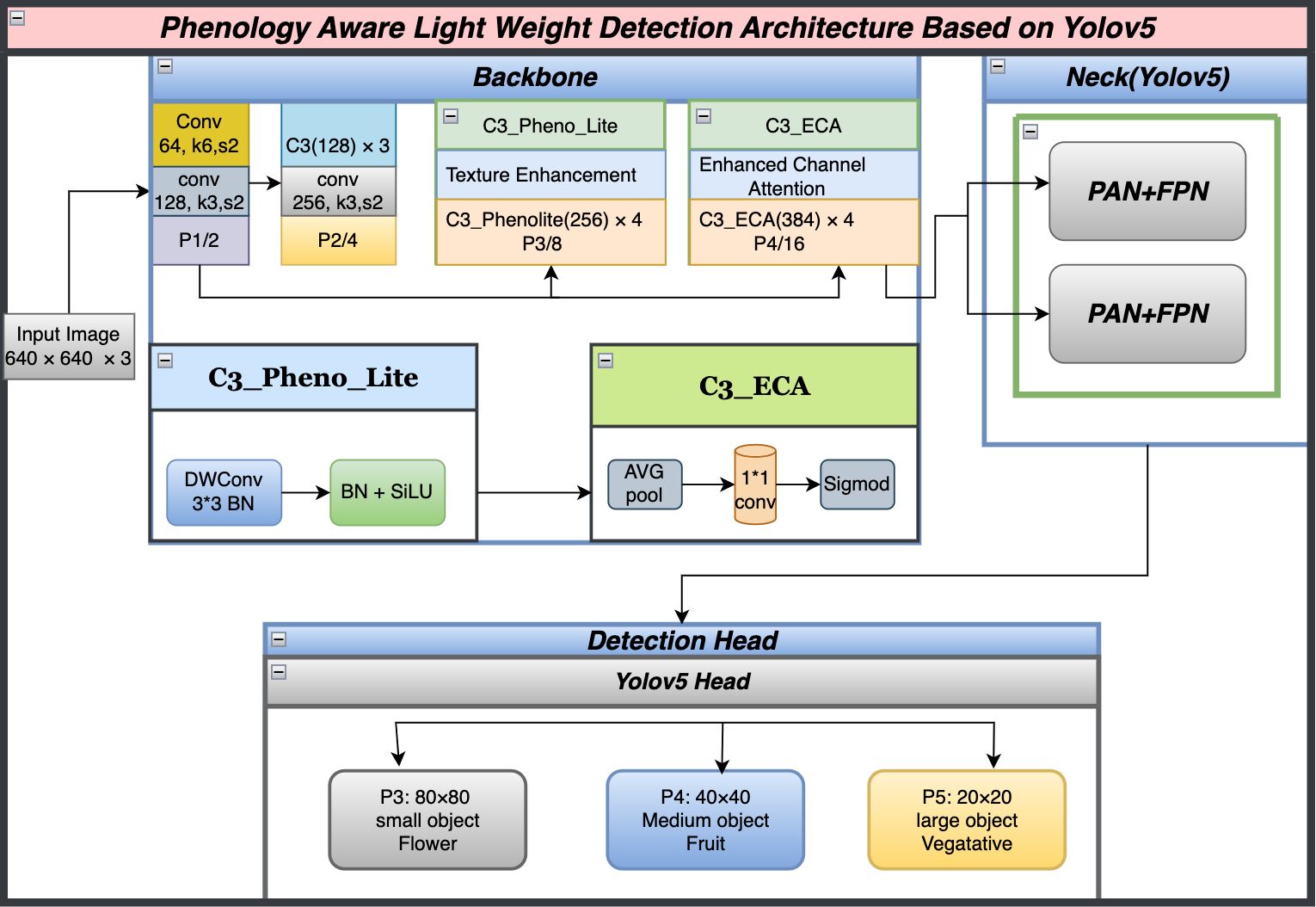}
\caption{Architecture of the proposed Pheno-Lite + ECA model based on YOLOv5s. The network consists of a Backbone, Neck, and Detection Head. Conv denotes convolution lay- ers where the first value (e.g., 64) indicates the number of output channels, followed by kernel size (k) and stride (s) (e.g., k6, s2). C3 represents the CSP-based bottleneck block. C3 PhenoLite enhances phenological texture features using DWConv, Batch Normalization (BN), and SiLU activation. C3 ECA integrates the Efficient Channel Attention (ECA) mechanism for channel-wise feature refinement. The PAN– FPN neck performs multi-scale feature fusion. P denotes the feature pyramid level used for detection: P3 (80×80) for small objects, P4 (40×40) for medium objects, and P5 (20×20) for large objects.}
\label{fig:arch}
\end{figure}

The proposed Pheno-Lite + ECA architecture as shown in Fig. \ref{fig:arch}, introduces targeted backbone refinements to the YOLOv5 framework to improve phenological stage discrimination under computational and deployment constraints. The design prioritizes efficient feature representation rather than exhaustive architectural expansion, reflecting the practical requirements of resource-limited greenhouse environments.
\subsubsection{Pheno-Lite Module}
The proposed Pheno-Lite module is a lightweight, phenology-aware backbone enhancement de- rived from the standard C3 block of YOLOv5s, designed to improve fine-grained spatial and texture feature representation for tomato growth stage recognition. The module introduces a depthwise residual refinement branch applied to the output of the C3 block, consisting of a depthwise convolution followed by batch normalization and SiLU activation, scaled by a residual factor before fusion with the main feature stream. This refinement selectively enhances localized spatial responses while preserving the original feature hierarchy and gradient stability. By leveraging depthwise convolution, the Pheno-Lite module increases sensitivity to subtle morphological cues such as leaf arrangement, flower emergence, and early fruit devel- opment with minimal parameter overhead, thereby maintaining computational efficiency and real-time inference capability in resource-constrained greenhouse environments.

\subsubsection{Efficient Channel Attention (ECA) Module}
To further improve discriminative feature learning at deeper semantic levels, the architecture integrates an Efficient Channel Attention (ECA) mechanism within selected C3 blocks. The ECA module models local cross-channel dependencies using global average pooling followed by a lightweight one-dimensional convolution, enabling adaptive channel-wise recalibration without dimensionality reduction\cite{Wang2020ECANet}. This design enhances inter- channel feature interaction while avoiding the complexity and information loss associated with fully connected attention mechanisms. Incorporated in a residual manner, the ECA module emphasizes phenology-relevant feature channels and suppresses redundant activations with negligible computational cost, contributing to improved robustness in distinguishing visually similar growth stages under varying illumination and background conditions.

\subsubsection{Neck and Detection Head}
The neck and detection head follow the standard YOLOv5s implementation without structural modification. A PANet style feature aggregation strategy is used to fuse multi-scale features via top-down and bottom-up pathways, combining high-resolution spatial information with deep semantic representations. The detection head operates at three output scales, corresponding to small, medium, and large receptive fields, and predicts bounding box coordinates, objectness confidence, and growth stage class probabilities. By preserving the original YOLOv5s neck and head design, the proposed architecture maintains reliable multi-scale detection performance while ensuring compatibil- ity with real-time inference requirements.

\subsubsection{Computational Efficiency}
Computational efficiency is a critical factor for real-time deployment in resource-constrained greenhouse environments. In this work, efficiency is evaluated in terms of the number of model parameters and the computational complexity measured by GFLOPs (Giga Floating-Point Operations). GFLOPs represent the total number of floating-point operations required to process a single input image during a forward pass of the network and provide an architecture-level measure of computational cost that is independent of hardware specifications. The GFLOPs of a convolutional neural network can be formulated as,\par
\begin{equation*}
\text{GFLOPs} = \frac{1}{10^9} \sum_{l=1}^{L}
\left(
2 \times C_l^{\mathrm{in}} \times C_l^{\mathrm{out}} \times K_l^2 \times H_l \times W_l
\right)
\end{equation*}
where \(C_l^{\mathrm{in}}\) and \(C_l^{\mathrm{out}}\) denote the input and output channels of layer \(l\), \(K_l\) is the convolution kernel size, and \(H_l \times W_l\) represents the spatial resolution of the corresponding feature map, in proposed model an input image of resolution $H_0 \times W_0 = 640 \times 640$, the output feature-map resolution $H_l \times W_l$ at the detection layers satisfies, $80 \times 80, 40 \times 40, 20 \times 20$, corresponding to the P3, P4, and P5 feature levels with total strides of 8, 16, and 32, respectively.
\section{Results}
\subsection{Performance of the Proposed Pheno-Lite + ECA Model}
The proposed Pheno-Lite + ECA architecture was evaluated on the heldout test set to assess its effectiveness for tomato growth stage detection under realistic greenhouse conditions. The model achieved a precision of 90.6\%, recall of 88.8\%, and an mAP@50 of 92.6\%, demonstrating reliable localization and classification performance across vegetative, flowering, and fruiting stages. The improved precision reflects enhanced confidence in predicted detections, suggesting effective suppression of false positives. This behavior can be attributed to the combined effect of localized spatial refinement introduced by the Pheno-Lite module and adaptive channel recalibration provided by the ECA mechanism. While recall remains comparable to standard lightweight detectors, the balanced precision–recall characteristics indicate that the proposed de- sign primarily improves feature discrimination rather than expanding detection coverage, which is desirable for stage-specific monitoring tasks.
\subsection{Quantitative Performance Analysis}
 The Table~\ref{tab:performance}, presents the quantitative detection performance of the proposed model in terms of precision, recall, and mAP@50. The results confirm consistent detection accuracy across all growth stages, highlighting the ability of the model to distinguish visually similar phenological phases. The strong mAP@50 score indicates accurate localization of the bounding box and stable classification behavior, validating the effectiveness of the proposed architectural refinements under practical greenhouse conditions.

\begin{table}[b]
\centering
\caption{Detection performance of the proposed Pheno-Lite + ECA model}
\label{tab:performance}
\begin{tabular}{|l|c|c|c|}
\hline
Class & Precision (\%) & Recall (\%) & mAP@50 (\%) \\
\hline
Overall     & 90.6 & 88.8 & 92.6 \\
Vegetative  & 93.2 & 91.8 & 96.9 \\
Flowering   & 90.6 & 85.6 & 89.6 \\
Fruiting    & 88.1 & 89.4 & 91.3 \\
\hline
\end{tabular}
\end{table}
\subsection{Precision–Recall and Confidence-Based Analysis}
\begin{figure}[h]
    \centering
    \subfloat[Precision-Recall Curve of the Proposed Model]{%
        \includegraphics[width=0.9\columnwidth]{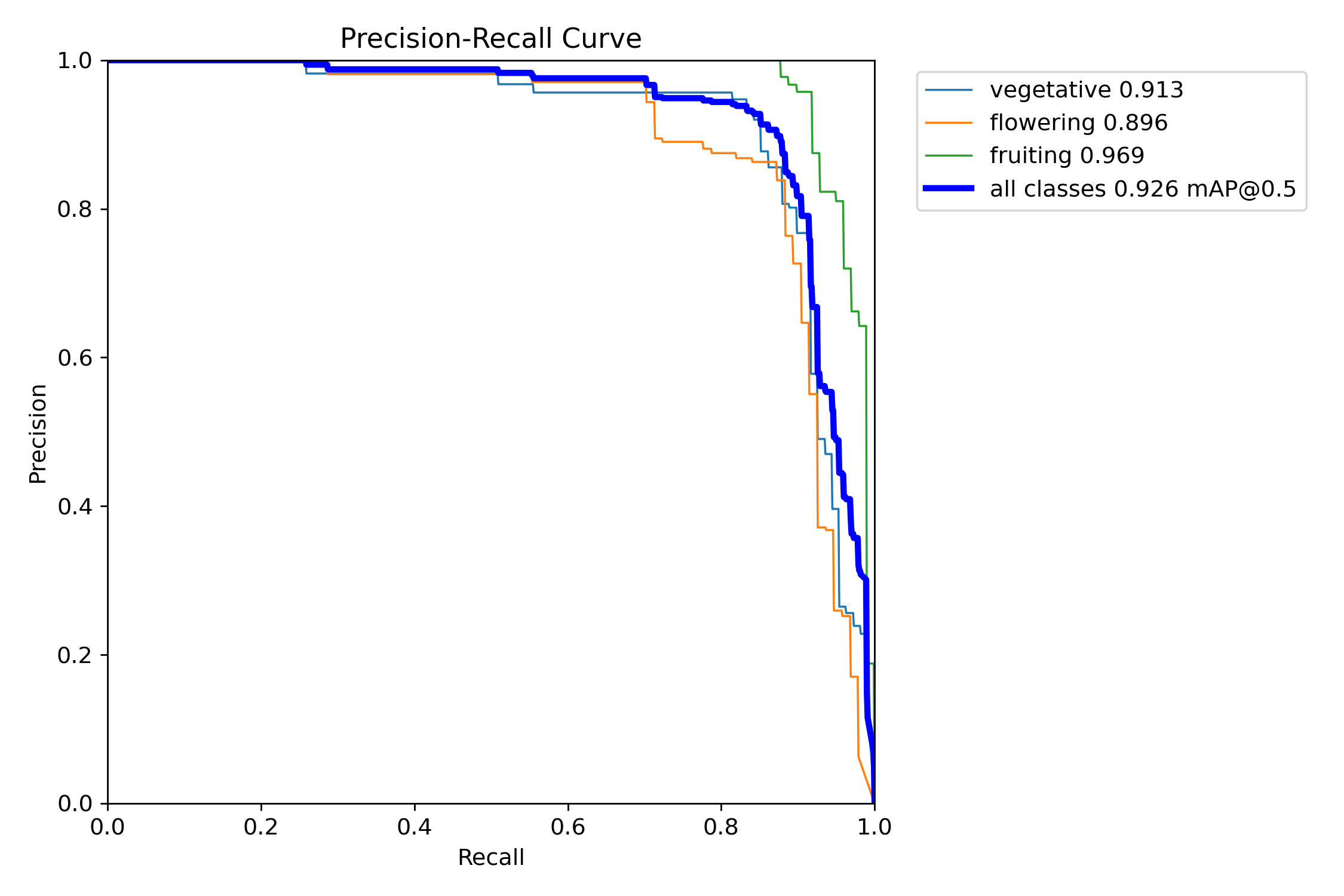}%
        \label{fig:vall}%
    }%
    \\[3pt]
    \subfloat[Precision-Confidence curve of the Proposed Model]{%
        \includegraphics[width=0.9\columnwidth]{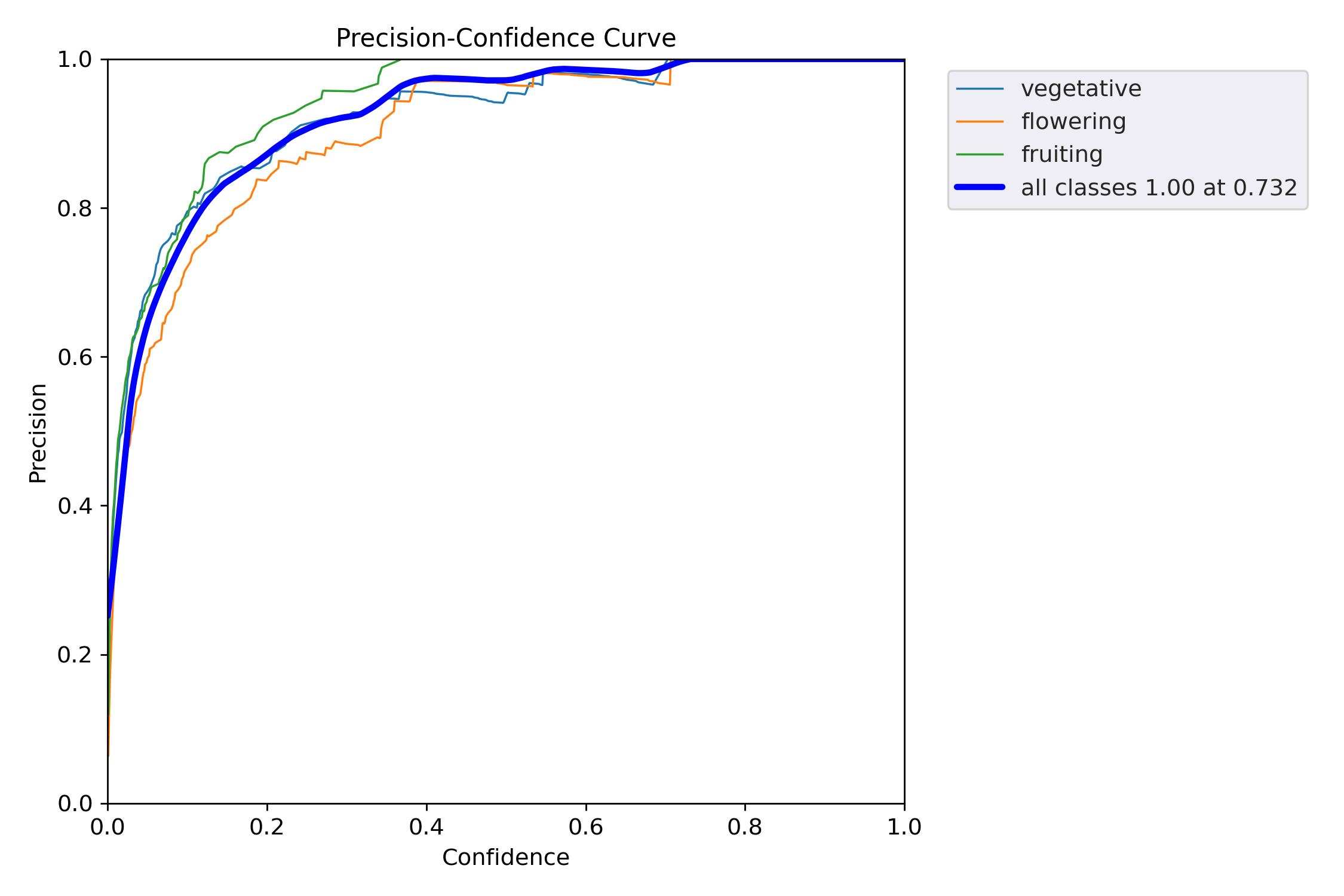}%
        \label{fig:valp}%
    }%
    \caption{Evaluation curves of the proposed Pheno-Lite + ECA model on the validation dataset. (a) Precision–recall (P–R) curve showing the trade-off between precision and recall at different confidence thresholds. (b) Precision–confidence curve illustrating how detection precision varies with the confidence threshold.}
    \label{fig:overallval}
\end{figure}

As shown in Fig. \ref{fig:vall}, the precision–recall curve illustrates stable detection performance across a broad recall range, indicating a favorable balance between detection sensitivity and reliability. High precision is maintained even as recall increases, reflecting robustness against false-positive detec- tions in cluttered scenes. The precision–confidence curve further demonstrates that higher confidence thresholds yield consistently reliable predictions, suggesting that the model produces well-calibrated confidence scores suitable for down- stream decision-making in automated greenhouse management systems as illustrated in Fig. \ref{fig:valp}.

\subsection{Validation Batch Labels and Predictions}
\begin{figure}[t]
\centering
\includegraphics[width=\columnwidth,height=0.35\textheight,keepaspectratio]{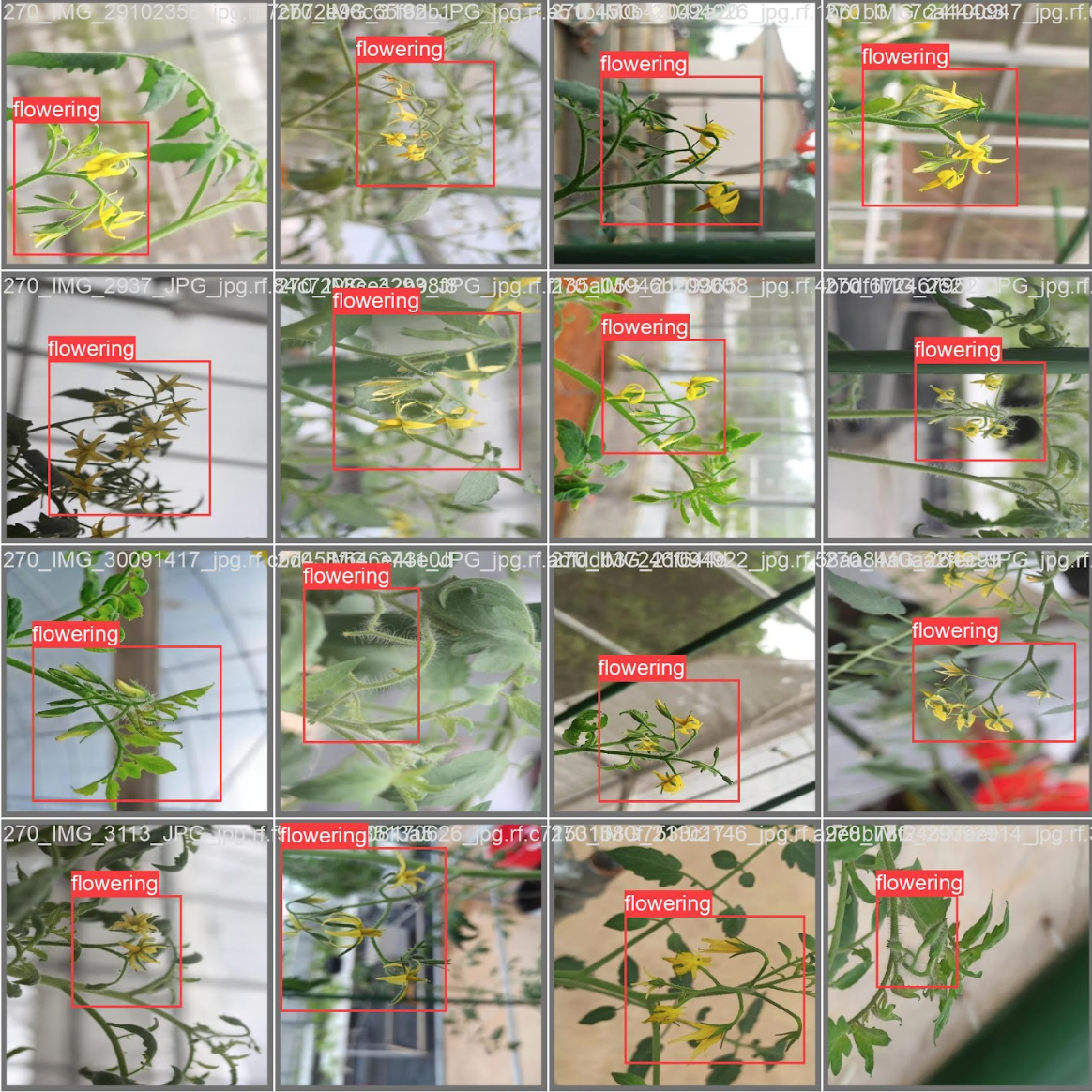}
\caption{Validation label batch1 of the proposed model. Shows the ground-truth annotation examples from the validation dataset, illustrating labeled tomato growth stages under greenhouse conditions.}
\label{fig:vlabel}
\end{figure}

Batch validation visualizations were used to qualitatively assess the integrity of the data set and the behavior of the model. The validation batch as shown in Fig.~\ref{fig:vlabel}, depicts the ground-truth annotations of the validation images, including the location of the bounding boxes and the corresponding class labels, and serve to verify the accuracy of the annotation, the spatial consistency and the representation of the class within the data set. \par
\begin{figure}[t]
\centering
\includegraphics[width=\columnwidth,height=0.35\textheight,keepaspectratio]{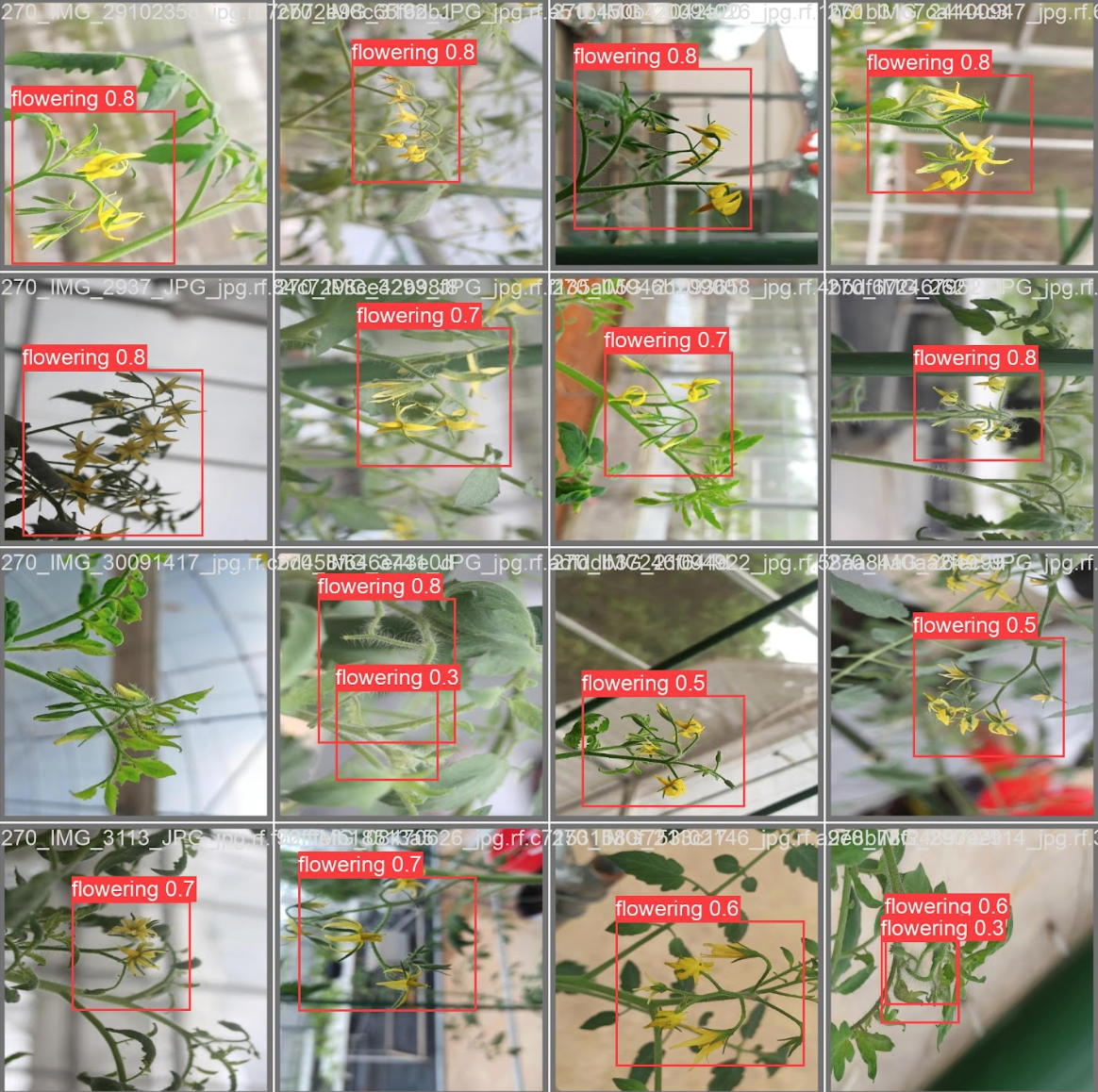}
\caption{Validation predict batch1 of the proposed model.Shows the detection results of the proposed model on validation images, representing the model’s inference output in terms of bounding boxes and class confidence scores.}
\label{fig:vpredict}
\end{figure}

The validation batch predictions represented in Fig.~\ref{fig:vpredict}, show the detection output generated by the trained model on the same validation images, showing the predicted bounding boxes, class assignments and confidence scores. Visual comparison between labels and predictions enables qualitative evaluation of localization accuracy, classification reliability, and common detection errors, where by complementing quantitative performance metrics and providing insight into the model’s behavior under practical greenhouse conditions.
\subsection{Model Complexity and Computational Efficiency}
The proposed Pheno-Lite + ECA model comprises approximately 4.0 million learnable parameters and requires about 10.9 billion floating-point operations to process a single image during one forward inference pass at an input resolution of 640 × 640 (GFLOPs 10.9). Compared to the YOLOv5s baseline, this represents a reduction in computational complexity, as reflected by the lower GFLOPs, while also achieving a more compact parameterization. The combination of a smaller parameter count and moderate computational cost enables efficient real-time inference, making the proposed model particularly well suited for deployment in resource-constrained greenhouse environments where both accuracy and efficiency are critical.
 
\subsection{Detection performance of baseline yolov5s}
The baseline YOLOv5s model was trained for 50 epochs under the same dataset, preprocessing, augmentation, and training configuration as the proposed framework to ensure a fair comparison. The model achieved an overall precision of 89.8\%, recall of 88.9\%, and mAP@50 of 93.6, indicating strong baseline performance for tomato growth stage detection in greenhouse environments as shown in Table~\ref{tab:baseline_performance},. Among the individual classes, the vegetative stage recorded the highest detection accuracy, with a precision of 94.4\% and mAP@50 of 96.7\%, reflecting its comparatively distinct visual features.\par
In contrast, the flowering and fruiting stages exhibited relatively lower performance, with mAP@50 values of 92.5\% and 91.6\%, respectively. This reduction can be attributed to increased inter-class visual similarity, partial occlusion, and complex background conditions commonly observed in greenhouse cultivation. While YOLOv5s demonstrates robust and computationally efficient detection capability, these results highlight its limited sensitivity to fine-grained phenological variations, thereby motivating the development of the proposed Pheno-Lite + ECA architecture for enhanced growth stage discrimination.
\begin{table}[b]
\centering
\caption{Detection performance of the baseline YOLOv5s model}
\label{tab:baseline_performance}
\begin{tabular}{|l|c|c|c|}
\hline
\textbf{Class} & \textbf{Precision (\%)} & \textbf{Recall (\%)} & \textbf{mAP@50 (\%)} \\
\hline
Overall     & 89.8 & 88.9 & 93.6 \\
Vegetative  & 94.4 & 91.8 & 96.7 \\
Flowering   & 88.0 & 88.9 & 92.5 \\
Fruiting    & 87.0 & 86.1 & 91.6 \\
\hline
\end{tabular}
\end{table}
\subsection{Precision and Recall curve }
\begin{figure}[b]
\centering
\includegraphics[width=\columnwidth,height=0.35\textheight,keepaspectratio]{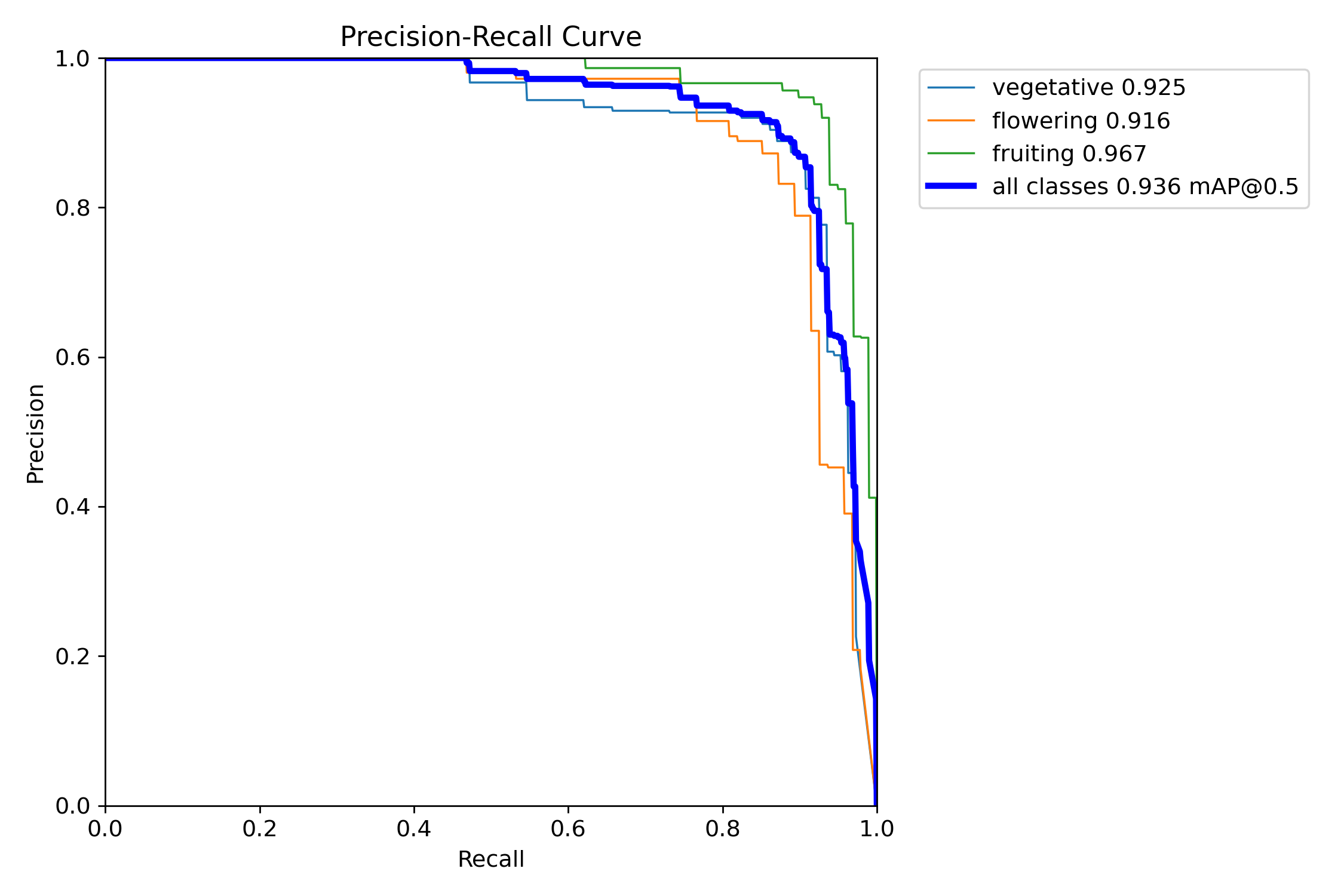}
\caption{Precision–recall (P–R) curve of the baseline YOLOv5s model, illustrating the relationship between precision and recall across different confidence thresholds on the validation dataset.}
\label{fig:precisiony}
\end{figure}

The precision–recall (P–R) curves of the baseline YOLOv5s model illustrate the trade-off between detection precision and recall across varying confidence thresholds. As depicted in Fig.~\ref{fig:precisiony}, curves demonstrate stable performance over a wide recall range, indicating consistent object localization and classification capability. Overall, the P–R analysis confirms the robustness of YOLOv5s as a baseline detector while revealing its limitations in capturing fine-grained phenological transitions, thereby justifying the need for architectural refinements.
\subsection{Model Complexity and Computational Efficiency of base model}
YOLOv5s serves as a compact and computationally efficient baseline for real-time object detection. When evaluated at an input resolution of 640 × 640, the model contains approximately 7 million trainable parameters and requires around 15.8 GFLOPs per forward pass. This architectural configuration enables effective multi-scale feature learning while maintaining relatively low inference latency compared to larger YOLO variants. However, the computational demand of YOLOv5s may still pose limitations for deployment in resource-constrained agricultural environments, such as greenhouses with limited edge-computing capabilities. As such, YOLOv5s provides a strong reference point for assessing the effectiveness of lightweight architectural modifications aimed at reducing model complexity while preserving detection performance.

\section{Discussion}
This study evaluates a lightweight, phenology-aware object detection framework tailored for tomato growth stage recognition in resource-constrained greenhouse environments. The proposed Pheno-Lite + ECA architecture is specifically designed to enhance discriminative learning of phenological features while simultaneously reducing model complexity and computational overhead. By prioritizing efficiency-aware architectural refinement over exhaustive network expansion, the framework targets practical deployment scenarios where computational resources, power availability, and automation infrastructure are limited. This design philosophy makes the proposed approach particularly well suited for greenhouse cultivation systems in Bhutan, where environmental variability and hardware constraints necessitate robust yet lightweight visual perception models.
Quantitative evaluation repersented in \ref{tab:comparison}, indicates that the baseline YOLOv5s attains a marginally higher overall mAP@50 of 93.6\% compared to 92.6\% achieved by the proposed Pheno-Lite + ECA model, reflecting the stronger raw representational capacity of the baseline architecture enabled by its larger parameter count and higher computational budget. In contrast, the proposed framework demonstrates superior overall precision (90.6\% versus 89.8\%) while maintaining comparable recall, indicating more confident and selective detections with reduced false-positive responses an important characteristic for growth stage–aware greenhouse manage- ment where incorrect stage identification can directly affect irrigation, fertilization, and microclimate control decisions. From a computational perspective, the proposed architecture offers a substantial reduction in model complexity, containing approximately 4.0 million parameters and requiring 10.9 GFLOPs at an input resolution of 640 × 640, compared to approximately 7 million parameters and 15.8 GFLOPs for the baseline YOLOv5s. This efficiency gain enables faster inference and lower memory consumption, which are critical for deployment on edge devices and low-power computing platforms commonly available in developing agricultural contexts. Importantly, this reduction in computational cost is achieved with only a marginal decrease in mAP@50, demonstrating an effective and practical trade-off between detection accuracy and computational efficiency. These characteristics directly align with the operational constraints of Bhutanese greenhouse systems, where limited automation, variable environmental conditions, and restricted computational infrastructure necessitate lightweight yet reliable phenology-aware detection models for real-world adoption.\par
To provide a balanced evaluation of detection accuracy and computational cost, the baseline YOLOv5s model and the proposed Pheno-Lite + ECA model are compared using grouped performance and efficiency metrics, visualized through a normalized radar chart. Performance metrics include Precision, Recall, and mAP@50, which respectively measure prediction correctness, object coverage, and localization accuracy based on the agreement between model predictions and ground-truth annotations. These metrics are grouped under performance because higher values directly indicate improved detection quality and are independent of model size, computational complexity, or hardware characteristics. In contrast, efficiency metrics capture the resource demands of the model and include the number of Parameters and GFLOPs. Parameters represent the total number of learnable weights in the network and reflect memory footprint and deployment feasibility, while GFLOPs quantify the total number of floating point operations required to process a single input image during one forward inference pass at a fixed resolution of 640 × 640, serving as an architecture-dependent indicator of computational complexity.\par
\begin{figure}[t]
\centering
\includegraphics[width=\columnwidth,height=0.35\textheight,keepaspectratio]{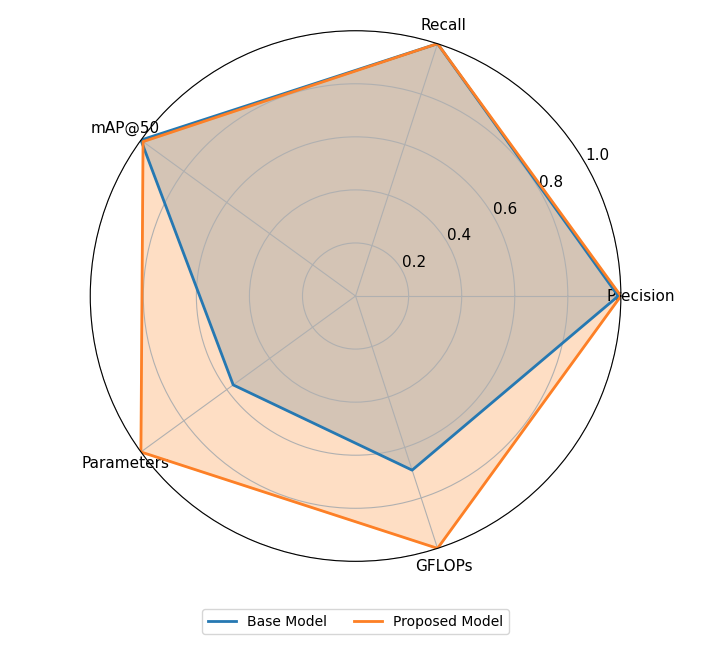}
\caption{Normalized radar chart depicting detection perfor- mance (Precision, Recall, and mAP@50) and computational efficiency (Parameters and GFLOPs) of YOLOv5s and the proposed Pheno-Lite + ECA model. Each axis represents a normalized metric, and the enclosed area indicates the overall performance–efficiency trade-off, with larger areas denoting a more favorable balance.}
\label{fig:normalized}
\end{figure}

Because the selected metrics differ in scale, units, and optimization direction, normalization to the range [0,1] is applied prior to visualization to enable fair and interpretable comparison. Performance metrics follow a higher-is-better criterion and are normalized relative to the best-performing model for each metric, ensuring that the comparison emphasizes relative detection capability rather than absolute values. Efficiency metrics follow a lower-is-better criterion and are normalized such that models with fewer parameters and lower computational cost obtain higher normalized scores. This unified normalization strategy converts all evaluation dimensions into a consistent higher-is-better representation, allowing them to be plotted on a common scale without bias. As illustrated  in Fig. \ref{fig:normalized}, the normalized radar chart enables intuitive visualization of multi-dimensional trade-offs between detection performance and computational efficiency. Based on this comparison, the proposed Pheno-Lite + ECA model demonstrates a more favorable overall balance than the YOLOv5s baseline, achieving comparable normalized performance across Precision, Recall, and mAP@50 while attaining substantially higher normalized efficiency scores due to its reduced parameter count and lower computational complexity.This balance makes the proposed framework particularly suitable for real-time deployment in resource-constrained greenhouse environments, where both accuracy and efficiency are critical for practical adoption.\par
Future work will further explore on improving current model through adaptive feature modeling to improve detection performance without significantly increasing computational overhead. Particular emphasis will be placed on collecting and curating time-series tomato growth data from Bhutanese greenhouse environments to capture temporal phenological transitions and seasonal variability, enabling more robust stage-aware modeling. Furthermore, coupling vision-based phenology detection with environmental sensor data (e.g., temperature, humidity, and soil moisture) which will enable holistic decision-support systems for automated irrigation and fertigation scheduling. Collectively, these extensions aim to advance the proposed framework toward scalable, context-aware, and fully deployable precision greenhouse systems suitable for Bhutan and similar resource-constrained agricultural settings.

\begin{table}[t]
\centering
\caption{Performance and Model Complexity Comparison Between Baseline YOLOv5s and Proposed Pheno-Lite + ECA}
\label{tab:comparison}
\renewcommand{\arraystretch}{1.2}
\begin{tabular}{|l|c|c|}
\hline
\multicolumn{3}{|c|}{\textbf{Detection Performance Comparison}} \\
\hline
\textbf{Metric} & \textbf{Base Model} & \textbf{Proposed Model} \\
\hline
Precision (\%) & 89.8 & 90.6 \\
Recall (\%)    & 88.9 & 88.8 \\
mAP@50 (\%)    & 93.6 & 92.6 \\
\hline
\multicolumn{3}{|c|}{\textbf{Model Complexity and Computational Efficiency}} \\
\hline
\textbf{Metric} & \textbf{Base Model} & \textbf{Proposed Model} \\
\hline
Parameters & $\sim$7M  & $\sim$4M \\
GFLOPs     & 15.8      & 10.9     \\
Input Size & $640 \times 640$ & $640 \times 640$ \\
\hline
\end{tabular}
\end{table}
\section{Conclusion}
This study presented Pheno-Lite + ECA, a lightweight and phenology-aware object detection framework for tomato growth stage recognition, specifically designed for greenhouse environments characterized by limited automation and constrained computational resources. Motivated by the agro-ecological and infrastructural conditions of Bhutan, where greenhouse cultivation is influenced by high-altitude climate variability, diffuse illumination, and manual management practices, the proposed architecture introduces targeted backbone refinements to enhance fine-grained phenological feature representation without incurring excessive computational cost. Experimental results demonstrate that the proposed model achieves high detection precision with a compact footprint of approximately 4.0 million parameters and 10.9 GFLOPs, offering an effective balance between accuracy and efficiency relative to the YOLOv5s baseline. These characteristics sup- port reliable real-time inference and practical deployment on low-power edge devices. Overall, this work provides a scalable and context-adaptive foundation for stage-specific greenhouse monitoring and contributes toward data-driven, climate-resilient precision agriculture in Bhutan and similar developing agricultural settings.
\section*{Acknowledgement}
The authors acknowledge the use of generative artificial intelligence tools (ChatGPT) based on GPT-5.3 during the preparation of this manuscript for language refinement and formatting assistance. The conceptualization, model design, experimental implementation, and analysis of results were carried out entirely by the authors. The authors take full responsibility for the accuracy, originality, and integrity of the work presented in this paper.
\bibliographystyle{IEEEtran}
\bibliography{references}
\end{document}